\documentclass[letterpaper]{article} 
\usepackage{aaai24}  
\usepackage{times}  
\usepackage{helvet}  
\usepackage{courier}  
\usepackage[hyphens]{url}  
\usepackage{graphicx} 
\usepackage{natbib}  
\usepackage{caption} 
\usepackage{algorithm}
\usepackage{algorithmic}

\usepackage{amsmath}
\usepackage{amssymb}
\usepackage{mathtools}
\usepackage{amsthm}
\usepackage{subfigure}
\usepackage{pdfpages}

\usepackage[capitalize,noabbrev]{cleveref}

\usepackage{algorithm}

\theoremstyle{plain}

\theoremstyle{definition}

\theoremstyle{remark}

\usepackage{newfloat}
\usepackage{listings}
\DeclareCaptionStyle{ruled}{labelfont=normalfont,labelsep=colon,strut=off} 
\floatstyle{ruled}
\newfloat{listing}{tb}{lst}{}
\floatname{listing}{Listing}
\title{Rating-Based Reinforcement Learning}
\author{
    Devin White\textsuperscript{\rm 1}, Mingkang Wu\textsuperscript{\rm 1},
    Ellen Novoseller\textsuperscript{\rm 2}, Vernon J. Lawhern\textsuperscript{\rm 2}, Nicholas Waytowich\textsuperscript{\rm 2},
    Yongcan Cao\textsuperscript{\rm 1}
}
\affiliations{
    \textsuperscript{\rm 1}University of Texas, San Antonio\\
    \textsuperscript{\rm 2}DEVCOM Army Research Laboratory

}

\usepackage{bibentry}

\begin{document}

\maketitle

\begin{abstract}
This paper develops a novel rating-based reinforcement learning (RbRL) approach that uses human ratings to obtain human guidance in reinforcement learning. Different from the existing preference-based and ranking-based reinforcement learning paradigms, based on human relative preferences over sample pairs, the proposed rating-based reinforcement learning approach is based on human evaluation of individual trajectories without relative comparisons between sample pairs. 
The rating-based reinforcement learning approach builds on a new prediction model for human ratings and a novel multi-class loss function. We finally conduct several experimental studies based on synthetic ratings and real human ratings to evaluate the performance of the new rating-based reinforcement learning approach. 
\end{abstract}

\section{Introduction} \label{sec:intro}
With the development of deep neural network theory and improvements in computing hardware, deep reinforcement learning (RL) has become capable of handling complex tasks with large state and/or action spaces (\textit{e.g.}, Go and Atari games) and yielding human-level or better-than-human-level performance~\cite{silver2016mastering,mnih2015human}. Numerous approaches, such as DQN~\cite{mnih2015human}, DDPG~\cite{lillicrap2015continuous}, PPO~\cite{schulman2017proximal}, and SAC~\cite{haarnoja2018soft} have been developed to address challenges such as stability, exploration, and convergence for various applications~\cite{li2019reinforcement} such as robotic control, autonomous driving, and gaming. Despite the important and fundamental advances behind these algorithms, one key obstacle for the wide application of deep RL is the required knowledge of a reward function, which is often unavailable in practical applications. 

Although human experts could design reward functions in some domains, the cost is high because human experts need to understand the relationship between the mission objective and state-action values and may need to spend extensive time adjusting reward parameters and trade-offs not to encounter adverse behaviors such as reward hacking~\cite{amodei2016concrete}. Another approach is to utilize qualitative human inputs \textit{indirectly} to learn a reward function, such that humans guide reward function design \textit{without} directly handcrafting the reward. 
Existing work on reward learning includes inverse reinforcement learning (IRL)~\cite{ziebart2008maximum}, preference-based reinforcement learning (PbRL)~\cite{christiano2017deep}, and the combination of demonstrations and relative preferences, e.g. learning from preferences over demonstrations~\cite{brown2019extrapolating}.

Existing human-guided reward learning approaches have demonstrated effective performance in various tasks. However, they suffer from some key limitations. For example, IRL requires expert demonstrations and hence, cannot be directly applied to tasks that are difficult for humans to demonstrate. 
PbRL is a practical approach to learning rewards for RL, since it is straightforward for humans to provide accurate relative preference information.
Yet, RL from pairwise preferences suffers from some key disadvantages. First, each pairwise preference provides only a single bit of information, which can result in sample inefficiency. 
In addition, due to their binary nature,
standard preference queries do not indicate how much better or worse one sample is than another. Furthermore, because preference queries are relative, they cannot directly provide a global view of each sample's absolute quality (good vs. bad); for instance, if all choices shown to the user are of poor quality, the user cannot say, ``A is better than B, but they're both bad!''. Thus, a PbRL algorithm may be more easily trapped in a local optimum, and cannot know to what extent its performance approaches the user's goal. Finally, PbRL methods often require strict preferences, such that comparisons between similar-quality or incomparable trajectories cannot be used in reward learning. While some works use weak preference queries~\cite{biyik2020asking, biyik2022aprel}, in which the user can state that two choices are equally preferable, there is no way to specify the quality (good vs. poor) of such trajectories; thus, valuable information remains untapped.  


The objective of this paper is to design a new rating-based RL (RbRL) approach that infers reward functions via multi-class human ratings. 
RbRL differs from IRL and PbRL in that it leverages human ratings on individual samples, whereas IRL uses demonstrations and PbRL uses relative pairwise comparisons.
In each query, RbRL displays one trajectory to a human and requests the human to provide a discrete rating. The number of rating classes can be as low as two, e.g. ``bad'' and ``good'', and can be as high as desired. For example, when the number of rating classes is $5$, the $5$ possible human ratings could correspond to ``very bad'', ``bad'', ``ok'', ``good'', and ``very good''. 
It is worth mentioning that the statement ``samples A and B are both rated as
`good' '' may provide more information than stating that ``A and B are equally preferable'', which can be inferred by the former.
However, ``A and B are equally preferable'' may be important
information for fine-tuning. In addition, a person can also intentionally assign high ratings to samples that contain rare states, which would be beneficial for addressing the exploration issue~\cite{ecoffet2019go} in RL. For both PbRL and RbRL, obtaining good samples requires exploration, and both will suffer
without any well-performing samples.

The main contributions of this paper are as follows. First, we propose a novel RbRL framework for reward function and policy learning from qualitative, absolute human evaluations. Second, we design a new multi-class cross-entropy loss function that accepts multi-class human ratings as the input. The new loss function is based on the computation of a relative episodic reward index 
and the design of a new multi-class probability distribution function based on this index.
Third, we conduct several experimental studies to quantify the impact of the number of rating classes on the performance of RbRL, and compare RbRL and PbRL under both synthetic and real human feedback. Our studies suggest that (1) too few or too many rating classes can be disadvantageous, (2) RbRL can outperform PbRL under both synthetic and real human feedback, and (3) people find RbRL to be less demanding, discouraging, and frustrating than PbRL.

\section{Related Work}

Inverse Reinforcement Learning (IRL) seeks to infer reward functions from demonstrations such that the learned reward functions generate behaviors that are similar to the demonstrations. Numerous IRL methods~\cite{ng2000algorithms}, such as maximum entropy IRL~\cite{ziebart2008maximum, wulfmeier2015maximum}, nonlinear IRL~\cite{finn2016guided}, Bayesian IRL~\cite{levine2011nonlinear, choi2011map, choi2012nonparametric}, adversarial IRL~\cite{fu2018learning}, and behavioral cloning IRL~\cite{szot2022bc} have been developed to infer reward functions. The need for demonstrations often makes these IRL methods costly, since human experts are needed to provide demonstrations.

Instead of requiring human demonstrations, PbRL~\cite{wirth2017survey,christiano2017deep, ibarz2018reward, liang2021reward,zhan2021human,xu2020preference, lee2021pebble, park2022surf} leverages human pairwise preferences over trajectory pairs to learn reward functions. Querying humans for pairwise preferences rather than demonstrations can dramatically save human time. In addition, by leveraging techniques such as adversarial neural networks~\cite{zhan2021human}, additional human time can be saved by learning a well-performing model to predict human preference. Another benefit of PbRL is that humans can provide preferences with respect to uncertainty to promote exploration~\cite{liang2021reward}. Despite these benefits, PbRL can be ineffective, especially for complex environments, because pairwise preferences only provide relative information rather than directly evaluating sample quality; while in some domains, sampled pairs may be selected carefully to infer global information, in practice, even if one sample is preferred over another, it does not necessarily mean that this sample is good. People can also have difficulty when comparing similar samples, thus taking more time and potentially yielding inaccurate preference labels. Notably, several works have sought to improve sample efficiency of PbRL; for instance, PEBBLE~\cite{lee2021pebble} considers off-policy PbRL, and SURF~\cite{park2022surf} explores data augmentations in PbRL. These contributions are orthogonal to ours, as they could straightforwardly be applied within our proposed RbRL framework.

Other methods for learning reward functions from humans include combining relative rankings and demonstrations, e.g. by inferring rewards via rankings over a pool of demonstrations~\cite{brown2020safe, brown2019extrapolating, brown2020better} to extrapolate better-than-demonstrator performance from the learned rewards, 
or first learning from demonstrations and then fine-tuning with preferences~\cite{ibarz2018reward, biyik2022learning}. Finally, in the TAMER framework~\cite{knox2009interactively, warnell2018deep, celemin2015coach}, a person gives positive (encouraging) and negative (discouraging) feedback to an agent with respect to specific states and actions, instead of over entire trajectories. These methods generally take actions greedily with respect to the learned reward, which may not yield an optimal policy in continuous control settings. 

\section{Problem Formulation}\label{sec:pre}

We consider a Markov decision process without reward (MDP\textbackslash R) augmented with ratings, which is a tuple of the form $(\mathcal{S}, \mathcal{A}, T, \rho, \gamma, n)$. Here, $\mathcal{S}$ is the set of states, $\mathcal{A}$ is the set of possible actions, $T: \mathcal{S} \times \mathcal{A} \times \mathcal{S} \to [0, 1]$ is a state transition probability function specifying the probability $p(s' \mid s, a)$ of reaching state $s' \in \mathcal{S}$ after taking action $a$ in state $s$, $\rho: \mathcal{S} \to [0, 1]$ specifies the initial state distribution, 
$\gamma$ is a discount factor, and $n$ is the number of rating classes. The learning agent interacts with the environment through rollout trajectories, where a length-$k$ trajectory segment takes the form $(s_1, a_1, s_2, a_2, \ldots, s_k, a_k)$. A \textit{policy} $\pi$ is a function that maps states to actions, such that $\pi(a \mid s)$ is the probability of taking action $a \in \mathcal{A}$ in state $s \in \mathcal{S}$.

In traditional RL, the environment would receive a reward signal $r: \mathcal{S} \times \mathcal{A} \to \mathbb{R}$, mapping state-action pairs to a numerical reward, such that at time-step $t$, 
the algorithm receives a reward $r_{t} = r(s_t, a_t)$, where $(s_t, a_t)$ is the state-action pair at time $t$.
Accordingly, the standard RL problem can be formulated as a search for the optimal policy $\pi^*$, where
$\pi^* = \arg\max_{\pi}\sum_{t=0}^{\infty} \mathbb{E}_{(s_t, a_t) \sim \rho_{\pi}} \Big [ \gamma^t r(s_t, a_t)\Big],
$
$ a_t \sim \pi (\cdot \vert  s_t) $, and $ \rho_{\pi} $ is the marginal state-action distribution induced by the policy $\pi$. Note that standard RL assumes the availability of the reward function $r$. When such a reward function is unavailable, standard RL and its variants may not be used to derive control policies. Instead, we assume that the user can assign any given trajectory segment $\tau = (s_1, a_1, \ldots, s_k, a_k)$ a rating in the set $\{0, 1, \ldots, n -1\}$ indicating the quality of that segment, where $0$ is the lowest possible rating, while $n - 1$ is the highest possible rating.

The algorithm presents a series of trajectory segments $\sigma$ to the human and receives corresponding human ratings. Let $X := \{(\sigma_i, c_i)\}_{i=1}^l$ be the dataset of observed human ratings, where $c_i \in \{0, \ldots, n -1\}$ is the rating class assigned to segment $\sigma_i$, and $l$ is the number of rated segments contained in $X$ at the given point during learning.

Note that descriptive labels can also be given to the rating classes. For example, for $n=4$ rating classes, we can call the rating class $0$ ``very bad'', the rating class $1$ ``bad'', the rating class $2$ ``good'', and the rating class $3$ ``very good''. With $n=3$ rating classes, we can call the rating class $0$ ``bad'', the rating class $1$ ``neutral'', and class $2$ ``good''.

\section{Rating-based Reinforcement Learning} \label{sec:RbRL}

Different from the binary-class reward learning in~\citet{christiano2017deep} that utilizes relative human preferences between segment pairs, RbRL utilizes non-binary multi-class ratings for individual segments. We call this a multi-class reinforcement learning approach based on ratings. 

\subsection{Modeling Reward and Return} Our approach learns a reward model $\hat{r}: \mathcal{S} \times \mathcal{A} \to \mathbb{R}$ that predicts state-action rewards $\hat{r}(s, a)$. We further define $\hat{R}(\sigma) := \sum_{t = 1}^{k} \gamma^{t - 1} \hat{r}(s_t, a_t)$ as the cumulative discounted reward, or the \textit{return}, of length-$k$ trajectory segment $\sigma$. Larger $\hat{R}(\sigma)$ corresponds to a higher predicted human rating for segment $\sigma$.
Next, we define $\tilde{R}(\sigma)$ as a function mapping a trajectory segment $\sigma$ to an estimated total discounted reward, normalized to fall in the interval $[0, 1]$ based on the dataset of rated trajectory segments $X$:
\begin{equation}\label{eq:eta}
    \tilde{R}(\sigma) = \frac{\hat{R}(\sigma) - \min_{\sigma' \in X} \hat{R}(\sigma')} {\max_{\sigma' \in X}\hat{R}(\sigma') - \min_{\sigma' \in X} \hat{R}(\sigma')}.
\end{equation} 

\subsection{Novel Rating-Based Cross-Entropy Loss Function}

To construct a new (cross-entropy) loss function that can take multi-class human ratings as the input, we need to estimate the human's rating class predictions. 
In addition, the range of the estimated rating class should belong to the interval $[0,1]$ for the cross-entropy computation. 
We here propose a new multi-class cross-entropy loss given by:
\begin{equation}\label{eq:crossentropy}
    L(\hat{r}) = -\sum_{\sigma\in X} \left( \sum_{i=0}^{n-1}\mu_{\sigma}(i) \log \big(Q_{\sigma}(i)\big) \right),
\end{equation}
where $X$ is the collected dataset of user-labeled segments, $\mu_{\sigma}(i)$ is an indicator that equals 1 when the user assigns rating $i$ to trajectory segment $\sigma$,
and $Q_{\sigma}(i) \in [0, 1]$ is the estimated probability that the human assigns the segment $\sigma$ to the $i$th rating class. Next, we will model the probabilities $Q_{\sigma}(i)$ of the human choosing each rating class. Notably, we do this \textit{without} comparing the segment $\sigma$ to other segments. 

\subsection{Modeling Human Rating Probabilities} \label{sec:Probability}

We next describe our model for $Q_{\sigma}(i)$ based on the normalized predicted returns $\tilde{R}(\sigma)$. To model the probability that $\sigma$ belongs to a particular class, we will first model separations between the rating classes in reward space.

We define rating class boundaries $\bar{R}_0, \bar{R}_1, \ldots, \bar{R}_n$ in the space of normalized trajectory returns such that $0 := \bar{R}_0 \le \bar{R}_1 \le \ldots \le \bar{R}_n := 1$.
Then, if a segment $\sigma$ has normalized predicted return $\tilde{R}(\sigma)$ such that $\bar{R}_i \le \tilde{R}(\sigma) \le \bar{R}_{i + 1}$, we wish to model that $\sigma$ belongs to rating class $i$ with the highest probability.

For example, when the total number of rating classes is $n = 4$, we aim to model the lower and upper return bounds for rating classes $0, 1, 2,$  and $3$, which for instance, could respectively correspond to ``very bad'', ``bad'', ``good'', and ``very good''. In this case, if $0 \le \tilde{R}(\sigma) < \bar{R}_1$, then we would like our model to predict that $\sigma$ most likely belongs to class 0 (``very bad''), while if $\bar{R}_2 \le \tilde{R}(\sigma) < \bar{R}_3$, then our model should predict that $\sigma$ most likely belongs to class 2 (``good'').


Given the rating category separations $\bar{R}_i$, we model $Q_{\sigma}(i)$ as a function of the normalized predicted returns $\tilde{R}(\sigma)$:
\begin{equation}\label{eq:Qfun_new}
     Q_{\sigma}(i) = \frac{e^{-k(\tilde{R}(\sigma)-\bar{R}_i)(\tilde{R}(\sigma)-\bar{R}_{i + 1})}}{\sum_{j=0}^{n-1} e^{-k(\tilde{R}(\sigma)-\bar{R}_j)(\tilde{R}(\sigma)-\bar{R}_{j + 1})}},
\end{equation}
where $k$ is a hyperparameter modeling human label noisiness, 
and the denominator ensures that $\sum_{i=0}^{n-1}Q_{\sigma}(i) = 1$, i.e. that the class probabilities sum to 1.

To gain intuition for Equation~\eqref{eq:Qfun_new}, note that when $\tilde{R}(\sigma) \in (\bar{R}_i, \bar{R}_{i + 1})$, such that the predicted return falls within rating class $i$'s predicted boundaries, then $-(\tilde{R}(\sigma)-\bar{R}_i)(\tilde{R}(\sigma)-\bar{R}_{i + 1})\ge 0$ while $-(\tilde{R}(\sigma)-\bar{R}_j)(\tilde{R}(\sigma)-\bar{R}_{j + 1})\le 0$ for all $j\neq i$. This means that $Q_{\sigma}(i)\ge Q_{\sigma}(j),~j\neq i$, so that the model assigns category $i$ the highest class probability, as desired. Furthermore, we note that $Q_{\sigma}(i)$ is maximized when $\tilde{R}(\sigma) = \frac{1}{2}(\bar{R}_i + \bar{R}_{i + 1})$, such that the predicted return falls directly in the center of category $i$'s predicted range. As $\tilde{R}(\sigma)$ becomes increasingly further from $\frac{1}{2}(\bar{R}_i + \bar{R}_{i + 1})$, the modeled probability $Q_\sigma(i)$ of class $i$ monotonically decreases. These probability trends are illustrated in Figure~\ref{fig:rating_class_probs} in the Appendix. 
We next show how to compute the 
class boundaries $\bar{R}_i,~i=1,\ldots,n - 1$.

\subsection{Modeling Boundaries between Rating Categories}
Next, we discuss how to model the boundaries between rating categories, $0 =: \bar{R}_0 \le \bar{R}_1 \le \ldots \le \bar{R}_n := 1$.
This requires selecting the range, or the upper and lower bounds of $\tilde{R}$, corresponding to each rating class.  
We determine these boundary values based on the distribution of $\tilde{R}(\sigma)$ for the trajectory segments $\sigma \in X$  and the number of observed samples in $X$ from each rating class. We select the $\bar{R}_i$ values such that 
the number of training data samples that the model assigns to each modeled rating class matches the number of samples in $X$ that the human assigned to that rating class.
Note that this does not require the predicted ratings based on $\tilde{R}(\sigma)$ to match the human ratings for $\sigma$ in the training data $X$, but 
ensures that the proportions of segments in the training dataset $X$ assigned to each rating class matches that in $X$. 
This matching in rating class proportions is desirable for learning an appropriate reward function based on human preference, since different humans could give ratings in significantly different proportions depending on their preferences and latent reward functions, as modeled by $\hat{R}$.

To define each $\bar{R}_i$ so that the number of samples in each modeled rating category reflects the numbers of ratings in the human data,
we first sort the estimated returns $\tilde{R}(\sigma)$ for all $\sigma \in X$ from lowest to highest, and label these sorted estimates as $\tilde{R}_1 \le \tilde{R}_2 \le \cdots\le \tilde{R}_l$, where $l$ is the cardinality of $X$. 
Denoting via $k_j$ the number of segments that the human assigned to rating class $j,~j \in \{0,\cdots,n-1\}$, we can then model each category boundary $\bar{R}_i, i \notin \{0, n\}$ (since $\bar{R}_0 := 0$ and $\bar{R}_n := 1$ by definition), as follows:
\begin{equation}  \bar{R}_i = \frac{\tilde{R}_{k_{i-1}^{\text{cum}}} + \tilde{R}_{1 + k_{i-1}^{\text{cum}}}}{2},\quad i \in \{1, 2, \ldots, n - 1\},
\end{equation}
where $k_i^{\text{cum}} := \sum_{j = 0}^i k_j$ is the total number of segments that the human assigned to any rating category $j \le i$.
When the user has not assigned any ratings within a particular category, i.e., $k_i = 0$ for some $i$, then we define the upper bound for category $i$ as $\bar{R}_{k_{i+1}} := \bar{R}_{k_{i}}$. 

This definition guarantees that when all normalized return predictions $\tilde{R}(\sigma), \sigma \in X,$ are distinct, then our model places $k_0$ segments within interval $[\bar{R}_0, \bar{R}_1)$, $k_i$ segments within each interval $(\bar{R}_i, \bar{R}_{i + 1})$ for $1 \le i \le n - 2$, and $k_{n-1}$ segments in $(\bar{R}_{n-1}, \bar{R}_n]$, and thus predicts that $k_i$ segments most likely have rating $i$.



\section{Synthetic Experiments}


\subsection{Setup} 

We conduct synthetic experiments based on the setup in~\citet{Lee2021BPref} to evaluate RbRL relative to the PbRL baseline~\cite{Lee2021BPref}. The code can be found at \url{https://rb.gy/tdpc4y}. The goal is to learn to perform a task by obtaining feedback from a teacher, in this case a synthetic human. 
For the PbRL baseline, we generate synthetic feedback such that in each queried pair of segments, the segment with the higher ground truth cumulative reward is preferred.
In contrast to the synthetic preferences between sample pairs in~\citet{Lee2021BPref}, RbRL was given synthetic ratings generated for individual samples, where these ratings were given by comparing the sample's ground truth return to the ground truth rating class boundaries. For simplicity, we selected these ground truth rating class boundaries so that rating classes are evenly spaced in reward space. 

For the synthetic PbRL experiments, we selected preference queries using the ensemble disagreement approach in~\citet{Lee2021BPref}. We extend this method to select rating queries for the synthetic RbRL experiments, designing an ensemble-based approach as in~\citet{Lee2021BPref} to select trajectory segments for which to obtain synthetic ratings. 
First, we train a reward predictor ensemble and obtain the predicted reward for every candidate segment and ensemble member.
We then select the segment with the largest standard deviation over the ensemble to receive a rating label. We study the Walker and Quadruped tasks in~\citet{Lee2021BPref}, with 1000 and 2000 synthetic queries, respectively. 

For all synthetic experiments, the reward network parameters are optimized to minimize the cross entropy loss~\eqref{eq:crossentropy} based on the respective batch of data via the computation of~\eqref{eq:Qfun_new}. 
We use the same neural network structures for both the reward predictor and control policy and the same hyperparameters as in~\citet{Lee2021BPref}.

\begin{figure}[htp]
     \centering
     \begin{subfigure}
         \centering
        \includegraphics[width=0.8\columnwidth]{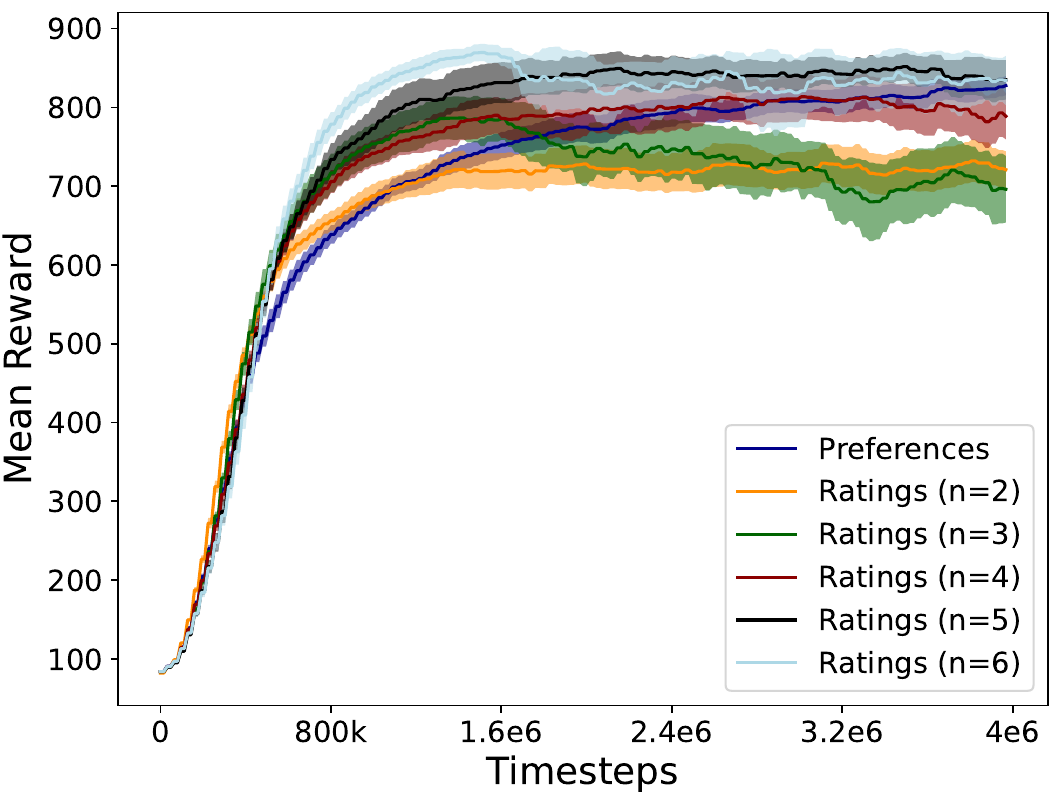}
         \label{fig:Walker}
     \end{subfigure} %
     \hfill
     \begin{subfigure}
         \centering
        \includegraphics[width=0.8\columnwidth]{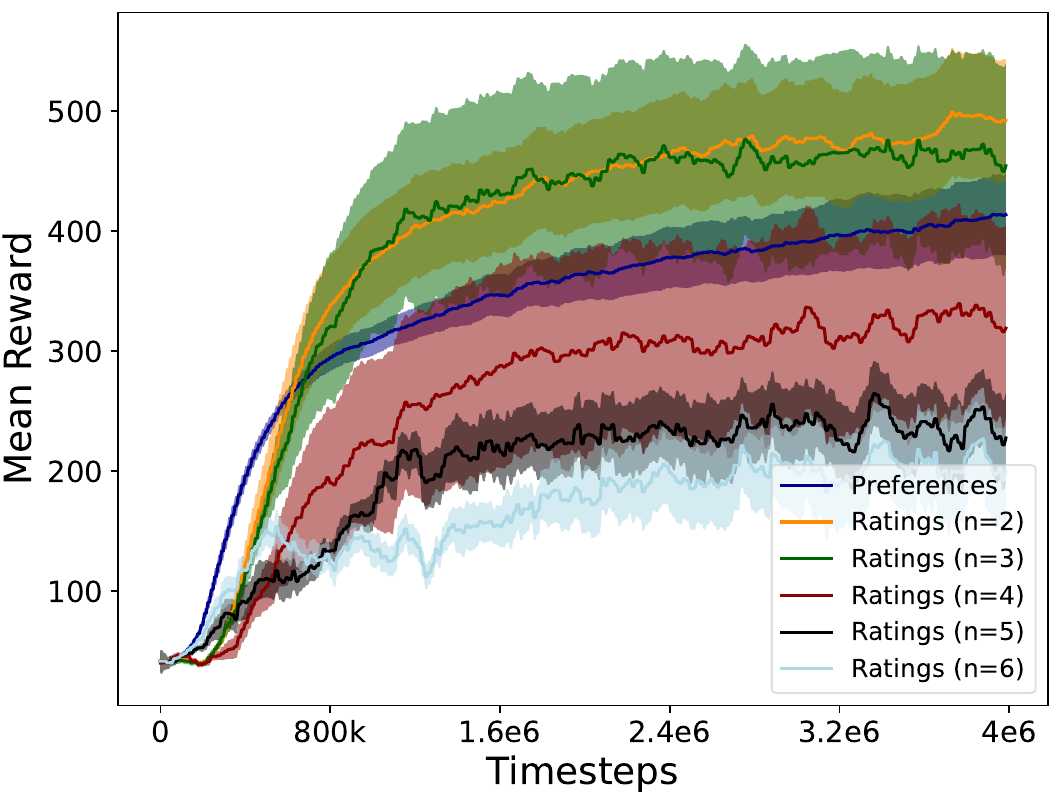}
         \label{fig:Quad}
     \end{subfigure}
\caption{Performance of RbRL in synthetic experiments for different $n$, compared to PbRL: mean reward $\pm$ standard error over 10 runs for Walker (top) and Quadruped (bottom).}
\label{fig:Bpref}
\end{figure}

\subsection{Results}

Figure \ref{fig:Bpref} shows the performance 
of RbRL for different numbers of rating classes (\textit{i.e.} values of $n$) and PbRL for two environments from~\citet{Lee2021BPref}: Walker and Quadruped. We observe that a higher number of rating classes yields better performance for  Walker. In addition, RbRL with $n=5,6$ outperforms PbRL. However, for Quadruped, while RbRL with $n=2,3$ still outperforms PbRL, a higher number of rating classes decreases performance; this decrease may be caused by the selection of rating class boundaries used to generate the synthetic feedback. The results indicate that RbRL is effective and can provide better performance than PbRL even if synthetic ratings feedback is generated using reward thresholds that are evenly distributed, without further optimization of their selection. We expect further optimization of the boundaries used to generate synthetic feedback to yield improved performance. 
For our experiments, we defined the rating boundaries by finding the maximum possible reward range for a segment and evenly dividing by the number of rating classes.


\section{Human Experiments}

\subsection{Setup} 
We conduct all human experiments by following a similar setup to~\citet{christiano2017deep}. In particular, our tests were approved by the UTSA IRB Office, including proper steps to ensure privacy and informed consent of all participants. In particular, the goal is to learn to perform a given task by obtaining feedback from a teacher, in this case a human. Different from PbRL in~\citet{christiano2017deep}, which asks humans to provide their preferences between sample pairs, typically in the form of short video segments, RbRL asks humans to evaluate individual samples, also in the form of short video segments, to provide their ratings, e.g., ``segment performance is good'' or ``segment performance is bad''. 

For all human experiments, 
we trained a reward predictor by minimizing the cross entropy loss~\eqref{eq:crossentropy} based on the respective batch of data via the computation of~\eqref{eq:Qfun_new}. 
We used the same neural network structures for both the reward predictor and control policy and the same hyperparameters as in~\citet{christiano2017deep}.

\subsection{RbRL with Different Numbers of Rating Classes}

\begin{figure}[ht]
\begin{center}
\centerline{\includegraphics[width=0.77\columnwidth]{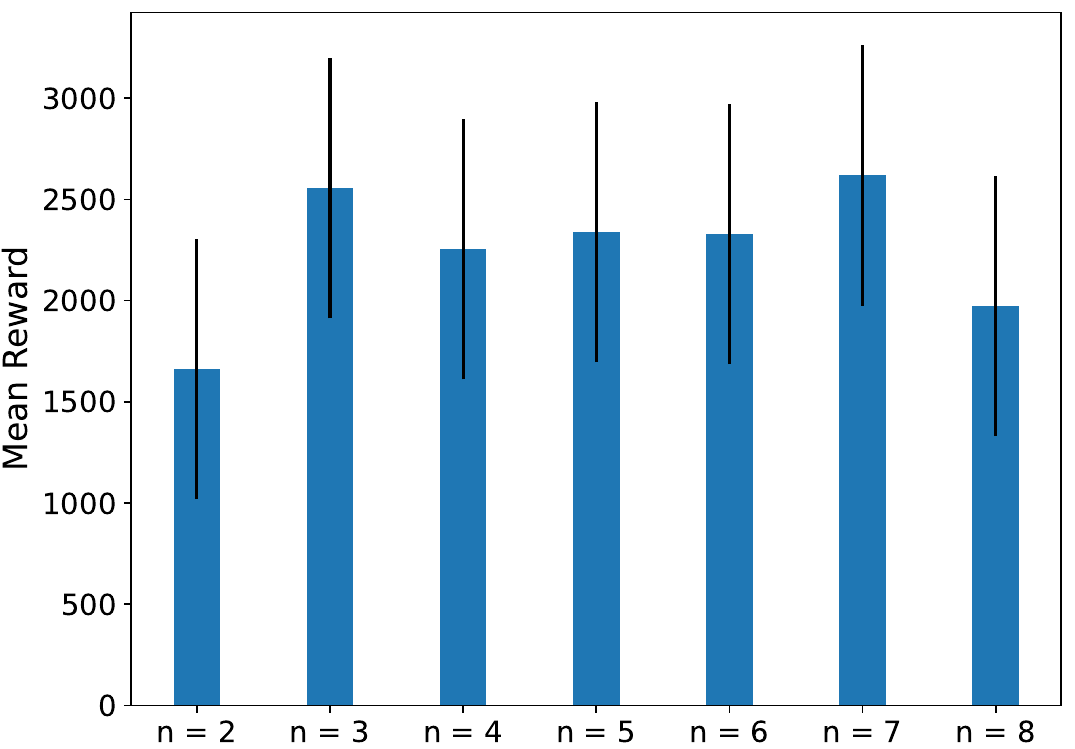}}
\caption{RbRL performance for different $n$ in a human experiment: performance in the Cheetah environment (mean $\pm$ standard error over 3 experiment runs). 
}
\label{fig:pvn}
\end{center}
\end{figure}

To evaluate the impact of the number of rating classes $n$ on RbRL's performance, we first conduct tests in which a human expert (an author on the study) provides ratings with $n=2,\ldots,8$ in the Cheetah MuJoCo environment. In particular, three experiment runs were conducted for each $n\in\{2,3,\ldots,8\}$. Fig.~\ref{fig:pvn} shows the performance of RbRL for each $n$.
It can be observed that RbRL performs better for $n\in\{3,4,\ldots,7\}$ than for $n \in \{2,8\}$, indicating that allowing more rating classes is typically beneficial. However, an overly large number of rating classes $n$ will lead to difficulties and inaccuracies in the human ratings, and hence $n$ must be set to a reasonable value. Indeed, for smaller $n$, one can more intuitively assign physical meanings to each $n$, whereas for overly large $n$, it becomes difficult to assign such physical meanings, and hence it will be more challenging for users to provide consistent ratings.

\subsection{RbRL Human User Study}

To evaluate the effectiveness of RbRL for non-expert users, we conducted an IRB-approved human user study.
We conducted tests on 3 of the OpenAI Gym MuJoCo Environments also used in~\citet{christiano2017deep}: Swimmer, Hopper and Cheetah. A total of 20 participants were recruited (7 for Cheetah, 7 for Swimmer, and 6 for Hopper). 
In our experiments, we provided a single 1 to 2 second long video segment to query users for each rating, while we provided pairs of 1 to 2 second videos to obtain human pairwise preferences. 
\textcolor{black}{For Cheetah, the goal is to move the agent to the right as fast as possible; this is the same goal encoded in the default hand-crafted environment reward. Similarly, the goal for Swimmer matches that of the default hand-crafted environment reward. However, for Hopper, we instructed users to teach the agent to perform a backflip, which differs from the goal encoded by the default hand-crafted environment reward. We chose to study the back flip task to see how well RbRL can learn new behaviors for which a reward is unknown.} Thus, the performance of Cheetah and Swimmer can be evaluated via the hand-crafted environment rewards, while the Hopper task cannot be evaluated via its hand-crafted environment reward. For Hopper, the performance of RbRL will be evaluated based on evaluating the agent's behavior when running the learned policies from RbRL. 

During the user study, each participant performed two tests---one for RbRL and one for PbRL---in one of the three MuJoCo environments, both for $n=2$ rating classes. To eliminate potential bias, we assigned each participant a randomized order in which to perform the PbRL and RbRL experiment runs. Because the participants had no prior knowledge of the MuJoCo environments tested, we provided sample videos to show desired and undesired behaviors so that the participants could better understand the task. 
Upon request, the participants could also conduct mock tests before we initiated human data collection. 
\textcolor{black}{For each experiment run, the participant was given 30 minutes to give rating/preference labels. Once finished, the participant filled out a questionnaire about the tested algorithm. The participant was then given a 10 minute break before conducting the second test and completing the questionnaire about the other algorithm. Afterwards, the participant completed a questionnaire comparing the two algorithms.} The questionnaires can be found in the Appendix. 
Policy and reward learning occurred during the 30 minutes in which the user answered queries, and then continued after the human stepped away until code execution reached 4 million environment time-steps. 

\subsubsection{Performance} \label{sec:performance}

\begin{figure}[ht]
     \centering
     \begin{subfigure}
         \centering
        \includegraphics[width=0.77\columnwidth]{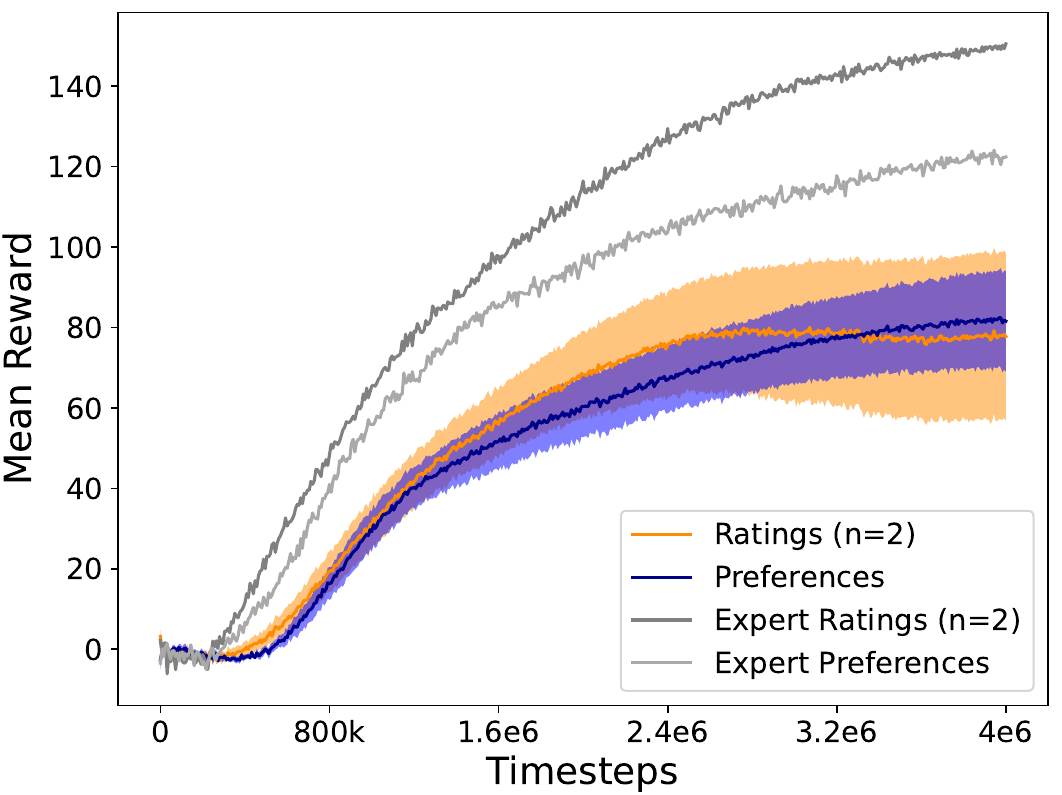}
         \label{fig:CheetahMean}
     \end{subfigure} %
     \hfill
     \begin{subfigure}
         \centering
        \includegraphics[width=0.77\columnwidth]{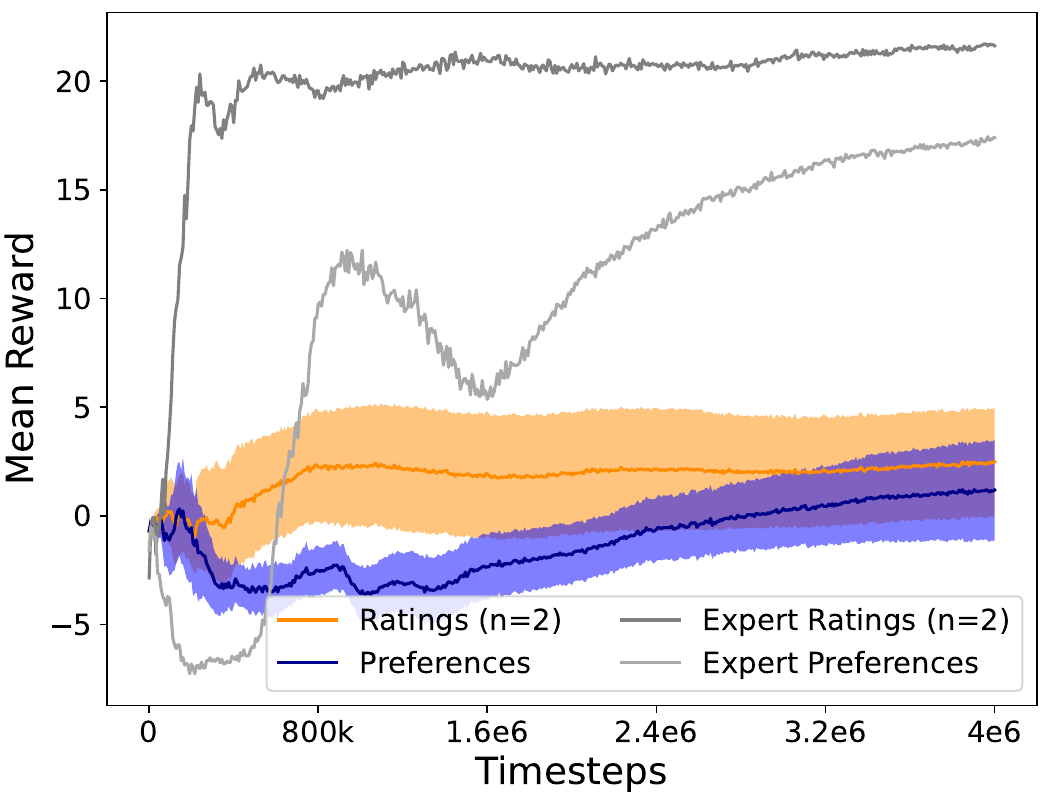}
         \label{fig:SwimmerMean}
     \end{subfigure}
\caption{Performance of RbRL and PbRL in the human user study: Cheetah (top) and Swimmer (bottom). For non-expert users, the plots show mean $\pm$ standard error over 7 users. The expert results are each over a single experiment run.}
\label{fig:meanreward}
\end{figure}

Figure~\ref{fig:meanreward} shows the performance of PbRL and RbRL 
across the seven participants for the Cheetah and Swimmer tasks. We see that RbRL performs similarly to or better than PbRL. In particular, RbRL can learn quickly in both cases, evidenced by the fast reward growth early during learning. Figure~\ref{fig:meanreward} additionally displays results when an expert (an author on the study) provided ratings and preferences for Cheetah and Swimmer.
For consistency, the same expert tested PbRL and RbRL in each environment. We observe that for the expert trials, RbRL performs consistently better than PbRL given the same human time. These results suggest that RbRL can outperform PbRL regardless of the user's environment domain knowledge. It can also be observed that the RbRL and PbRL trials with expert users outperform the trials in which feedback is given by non-experts.

\begin{figure}[ht]
    \centering
         \begin{subfigure}
         \centering
        \includegraphics[width=0.77\columnwidth]{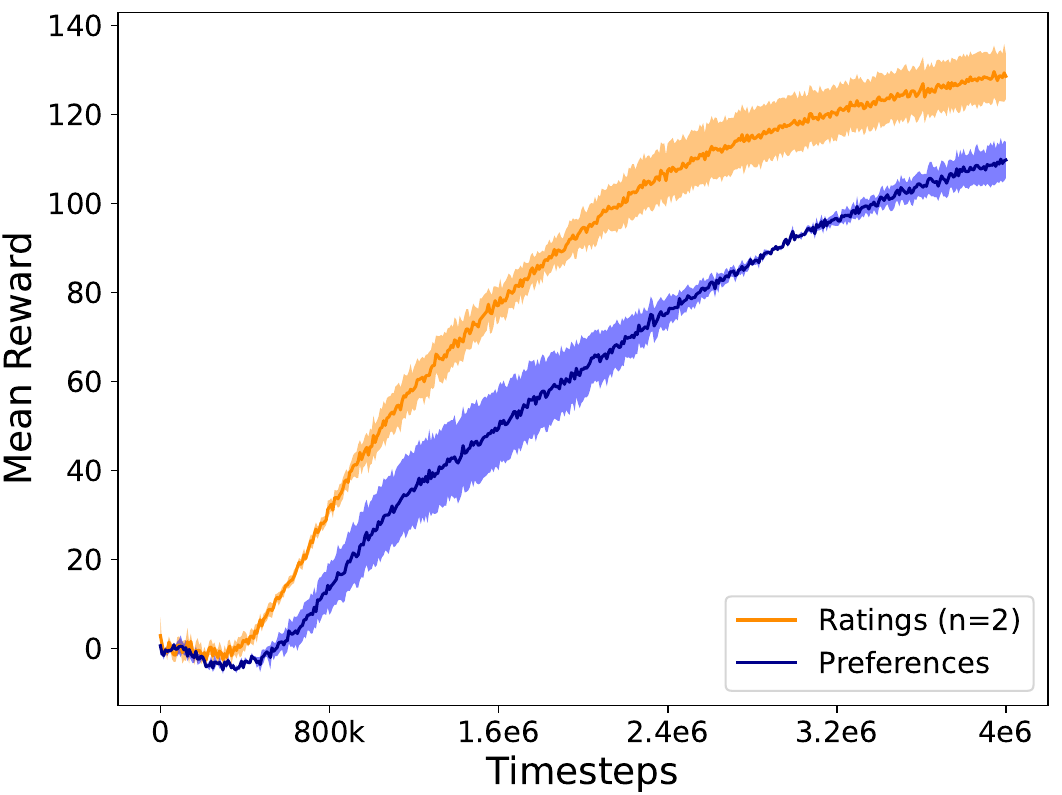}
         \label{fig:T3CheetahMean}
     \end{subfigure} %
     \hfill
    \begin{subfigure}
         \centering
        \includegraphics[width=0.77\columnwidth]{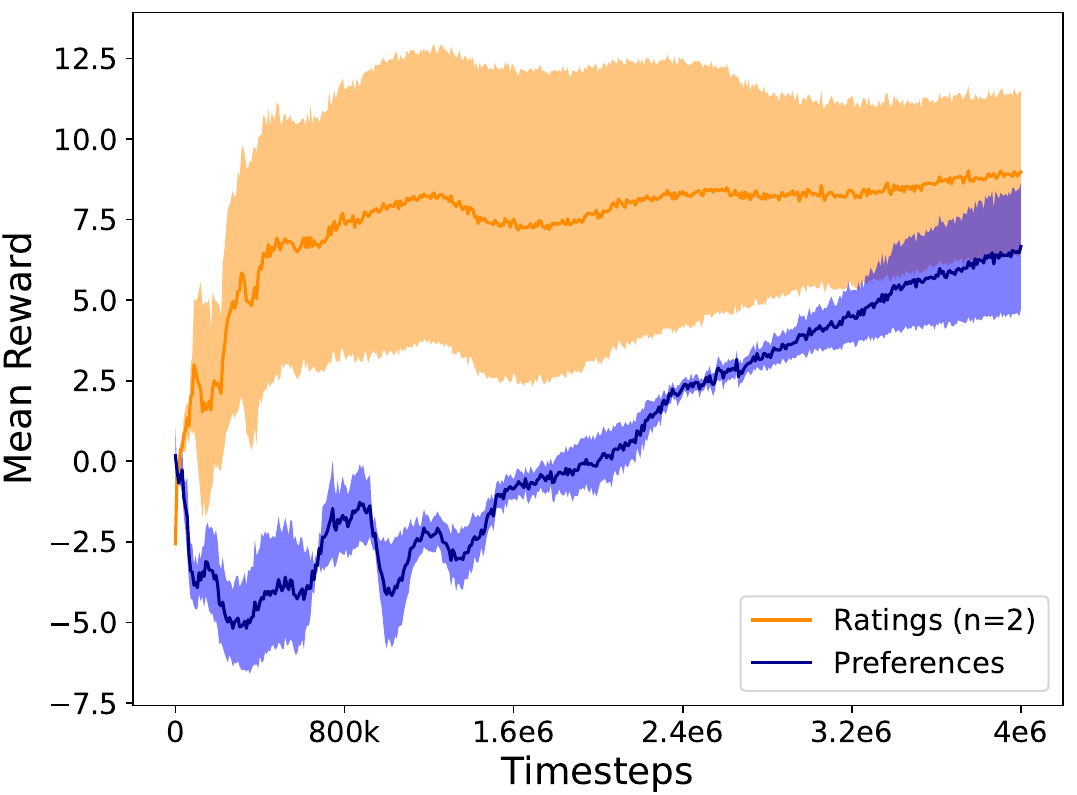}
         \label{fig:T3SwimmerMean}
     \end{subfigure}
\caption{RbRL and PbRL performance for the top 3 (non-expert) user study participants: mean reward $\pm$ standard error over the 3 experiment runs each for Cheetah (top) and Swimmer (bottom).}
\label{fig:top3}
\end{figure}



Although RbRL performs similarly to PbRL in the Cheetah task, we observed that some participants performed very poorly in this environment, perhaps due to lack of understanding of the task. In the Appendix, the raw data of all participants for Cheetah and Swimmer is provided to show performance under each individual participant's guidance. From the individual results for Cheetah (RbRL), we can see that one of the trials performs very poorly (with the final reward less than $-10$). For all other tests, including both PbRL and RbRL, the final reward is in positive territory, usually more than $20$. 
Hence, it may be more meaningful to evaluate the mean results for individuals who perform reasonably. 
Figure~\ref{fig:top3} shows the mean reward for the top 3 non-expert users at different iterations for Cheetah and Swimmer. It can be observed that RbRL consistently outperforms PbRL and learns the goal faster than PbRL. 

To compare PbRL and RbRL in the Hopper backflip task, we ran the learned policies for the 6 participants to generate videos. Videos for the best learned policies from PbRL and RbRL can be found at \url{rb.gy/nt1qm6}, and indicate that (1) both RbRL and PbRL can learn the backflip, and (2) the backflip learned via RbRL fits better with our understanding of a desired backflip. 

\subsubsection{User Questionnaire Results}

\begin{figure}[hhhhh]
    \centering
         \begin{subfigure}
         \centering
        \includegraphics[width=0.75\columnwidth]{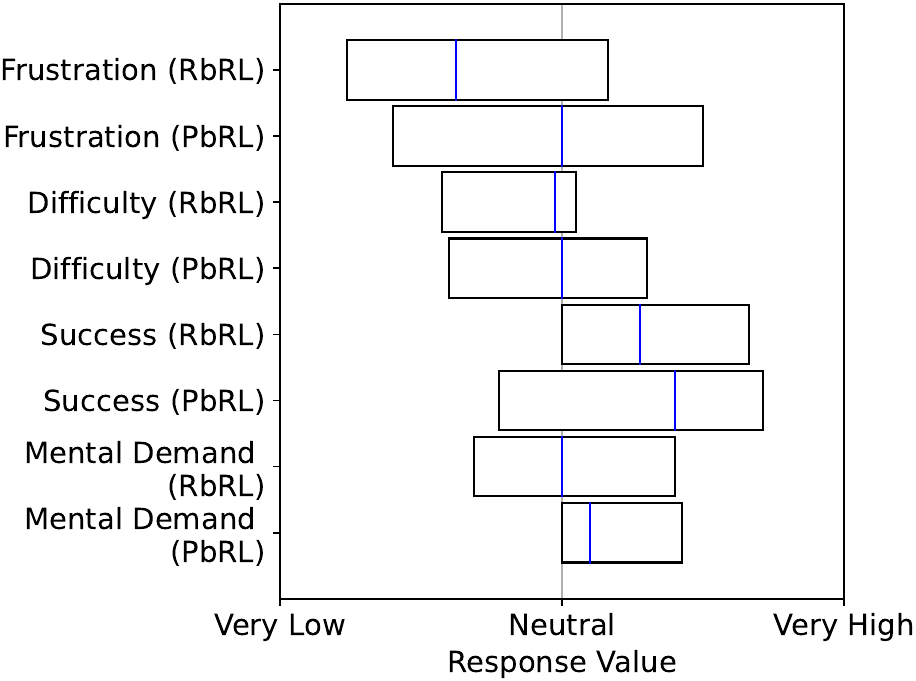}
         \label{fig:response}
     \end{subfigure} %
     \hfill
    \begin{subfigure}
         \centering
        \includegraphics[width=0.7\columnwidth]{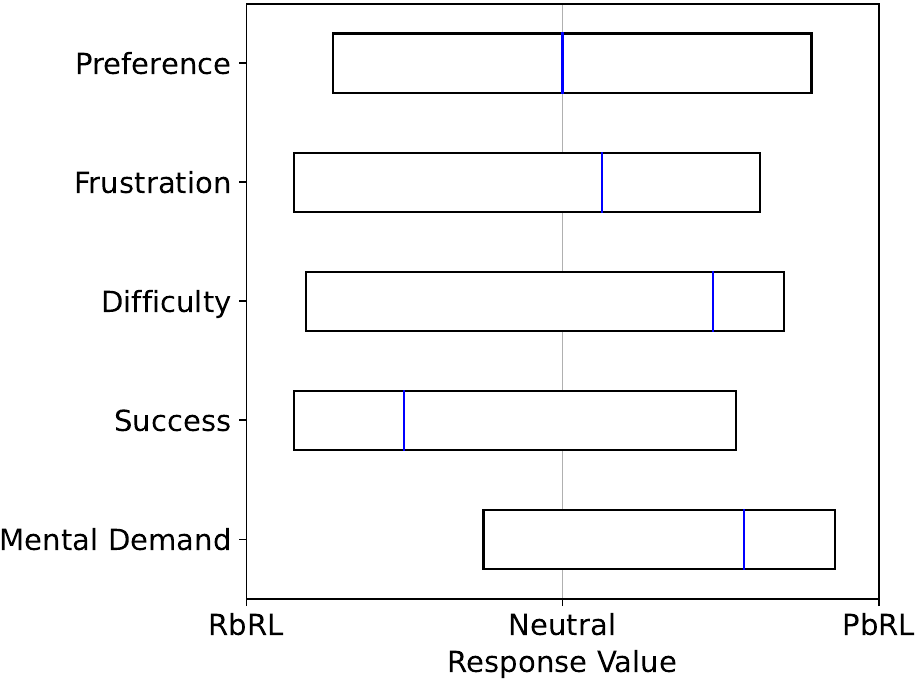}
         \label{fig:comparison}
     \end{subfigure}
\caption{Participants' responses to survey questions about RbRL and PbRL. The set of survey questions is detailed in the Appendix.
\textcolor{black}{The blue bar indicates the median and the edges depict the 1st quartile (left) and 3rd quartile (right).}
}
\label{fig:survey}
\end{figure}

The previous results in this section focus on evaluating the performance of RbRL and PbRL via the ground-truth environment reward (Cheetah and Swimmer) and the learned behavior (Hopper). 
To understand how the non-expert users view their experience of giving rating and preferences, we conduct a post-experiment user questionnaire,
shown in the last section of the Appendix. The questionnaire asked users for feedback about their experience supervising PbRL and RbRL and to compare PbRL and RbRL. 
Figure~\ref{fig:survey} displays the normalized survey results from the 20 user study participants.
In particular, the top subfigure of Figure~\ref{fig:survey} shows the participants' responses with respect to their separate opinions about PbRL and RbRL. These responses suggest that PbRL is more demanding and difficult than RbRL, leading users to feel more insecure and discouraged than when using RbRL. The bottom subfigure of Figure~\ref{fig:survey} shows the survey responses when users were asked to compare PbRL and RbRL; these results confirm the above findings and also show that users perceive themselves as completing the task more successfully when providing ratings (RbRL). One interesting observation is that the participants prefer RbRL and PbRL equally, which differs from the other findings. However, one participant stated that he/she preferred PbRL because PbRL is more challenging, which is counter-intuitive. This suggests that ``liking'' one algorithm more than the other is a very subjective concept, making the responses for this question less informative than those for the other survey questions.


\subsubsection{Human Time}~\label{sec:HumanTime}
\begin{figure}[hhhhh]
\begin{center}
\centerline{\includegraphics[width=0.7\columnwidth]{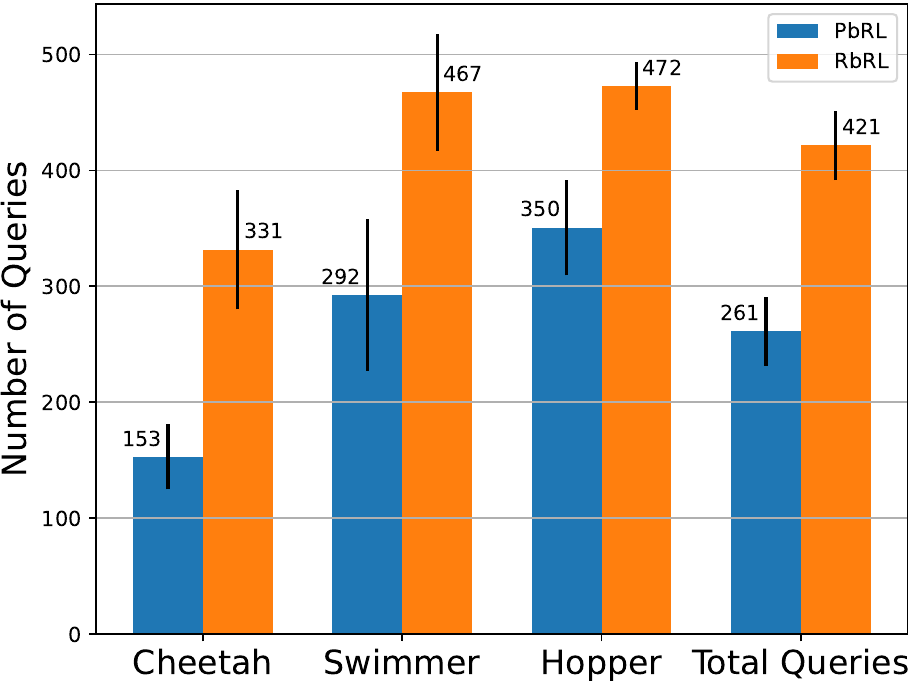}}
\caption{Number of queries provided in 30 minutes in our human user study (mean $\pm$ standard error).}
\label{fig:time}
\end{center}
\vskip -0.2in
\end{figure}

We also conducted a quantitative analysis of human time effectiveness when humans were asked to give ratings and preferences. Figure~\ref{fig:time} shows the average number of human queries provided in 30 minutes for Cheetah, Swimmer, Hopper, and for all three environments combined. 
It can be observed that the participants can provide more ratings than pairwise preferences in all environments, indicating that it is easier and more efficient to provide ratings than to provide pairwise preferences. On average, participants can provide approximately 14.03 ratings per minute, while they provide only 8.7 preferences per minute, which means that providing a preference requires 62\% more time than providing a rating. For Cheetah, providing a preference requires 100\%+ more time than providing a rating, which is mainly due to the need to compare video pairs that are very similar. For Swimmer and Hopper, the environments and goals are somewhat more complicated. Hence, providing ratings can be slightly more challenging, but is still easier than providing pairwise preferences.

\section{Discussion and Open Challenges}




One key difference between PbRL and RbRL is the value of the acquired human data. Because ratings in RbRL are not relative, they have the potential to provide more global value than preferences, especially when queries are not carefully selected. For environments with large state-action spaces, ratings can provide more value for reward learning. One limitation of ratings feedback is that the number of data samples in different rating classes can be very different, leading to imbalanced datasets. Reward learning in RbRL can be negatively impacted by this data imbalance issue (although our experiments still show the benefits of RbRL over PbRL). Hence, on-policy training with a large number of training steps may not help reward learning in RbRL because the collected human ratings data can become very unbalanced.
We expect that addressing the data imbalance issue would further improve RbRL performance.

One challenge for RbRL is that ratings may not be given consistently during learning, especially considering users' attention span and fatigue level over time. Future work includes developing mechanisms to quantify users' consistency levels, the impact of user inconsistency, or solutions to user inconsistency. 
Another potential limitation of RbRL is that it learns a less refined reward function than PbRL because RbRL does not seek to distinguish between samples from the same rating class. Hence, future work could integrate RbRL and PbRL to create a multi-phase learning strategy, where RbRL provides fast initial global learning while PbRL further refines performance via local queries based on sample pairs. 

One open challenge is the lack of effective human interfaces in existing code bases. For example, in~\citet{Lee2021BPref}, only synthetic human feedback is available. Although a human interface is available for the algorithm in~\citet{christiano2017deep}, the use of Google cloud makes it difficult to set up and operate efficiently. One of our future goals is to address this challenge by developing an effective human interface for reinforcement learning from human feedback, including preferences, ratings, and their variants. 

\section*{Acknowledgements}
The authors were supported in part by the Army Research Lab under grant W911NF2120232, Army Research Office under grant W911NF2110103, and Office of Naval Research under grant N000142212474. We thank Feng Tao, Van Ngo, Gabriella Forbis for their helpful feedback, code, and tests.



\bibliography{example_paper}

\newpage
\appendix
\onecolumn






\section{RbRL Implementation Details}

All tests and experiments of our RbRL and PbRL algorithms were trained for $4$ million time-steps with a segment length of $50$. For the Walker environment, 1,000 synthetic labels were provided with the reward predictor being updated every 20,000 time-steps. For the Quadruped environment, 2,000 synthetic labels were provided with the reward predictor being updated every 30,000 time-steps. All other hyper-parameters are the same as those in~\citet{Lee2021BPref}.

\section{Raw Data for Agent Performance from Individual Participants in the User Study}\label{sec:Raw}

\begin{figure}[ht]
     \centering
     \subfigure[Individual results for Cheetah (PbRL)]{
         \centering
         \includegraphics[width=0.475\columnwidth]{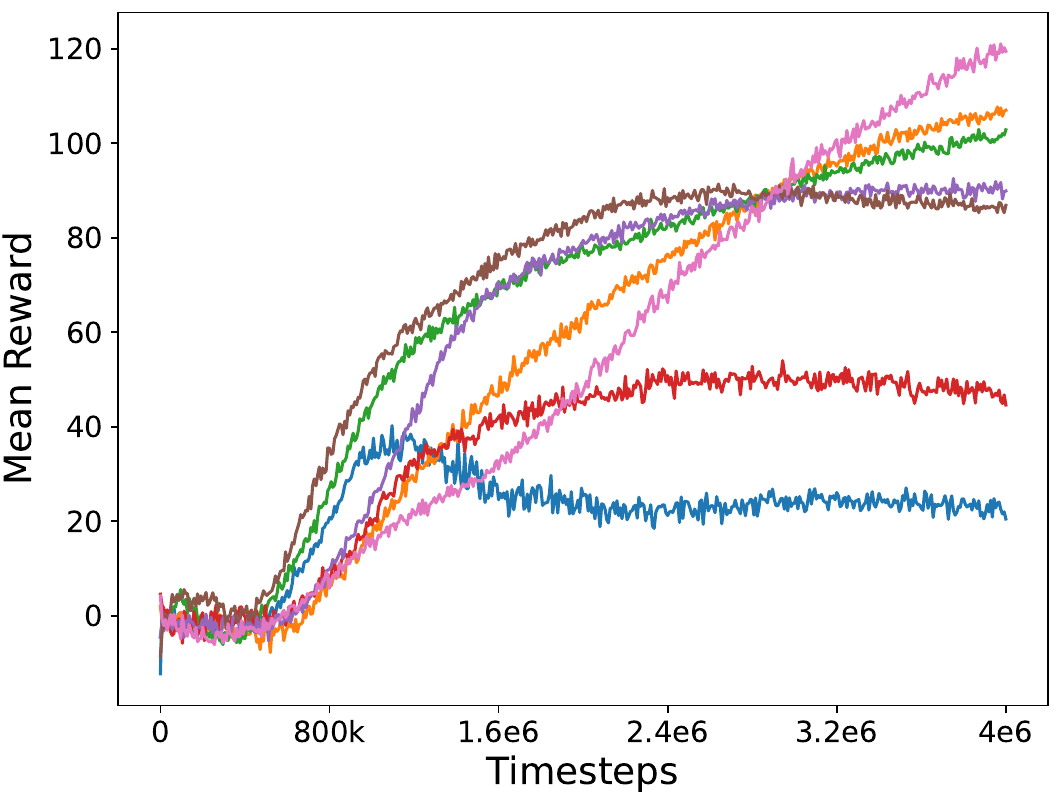}
         \label{fig:Cheetah_pbrl_individual}
     } %
     \hfill
     \subfigure[Individual results for Cheetah (RbRL with $n=2$)]{
         \centering
         \includegraphics[width=0.475\columnwidth]{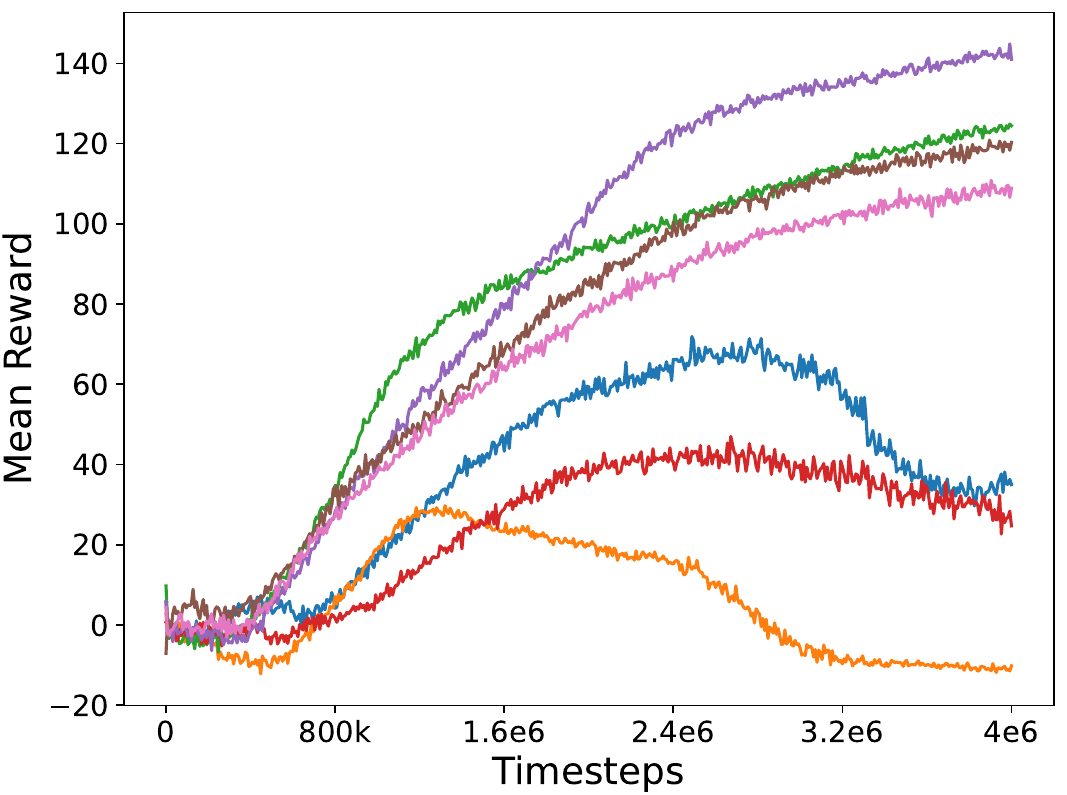}
         \label{fig:cheetah_rbrl_individual}
     }
     \subfigure[Individual results for Swimmer (PbRL)]{
         \centering
         \includegraphics[width=0.475\columnwidth]{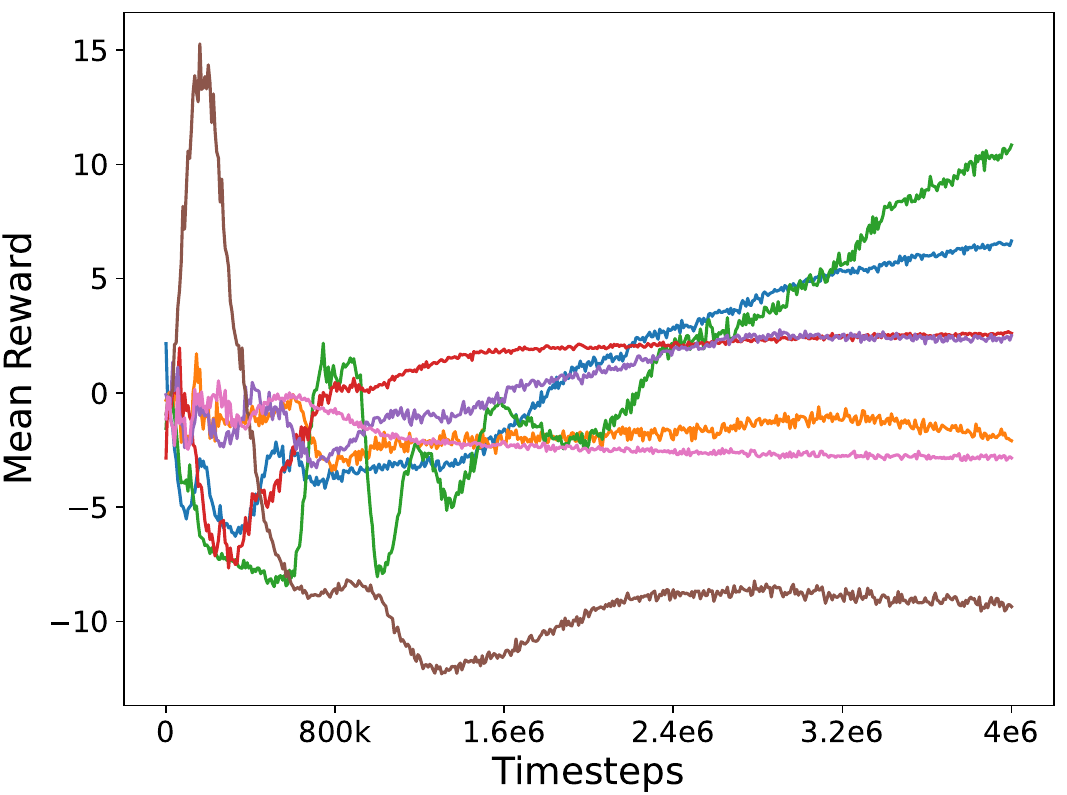}
         \label{fig:swimmer_pbrl_individual}
     } %
     \hfill
     \subfigure[Individual results for Swimmer (RbRL with $n=2$)]{
         \centering
         \includegraphics[width=0.475\columnwidth]{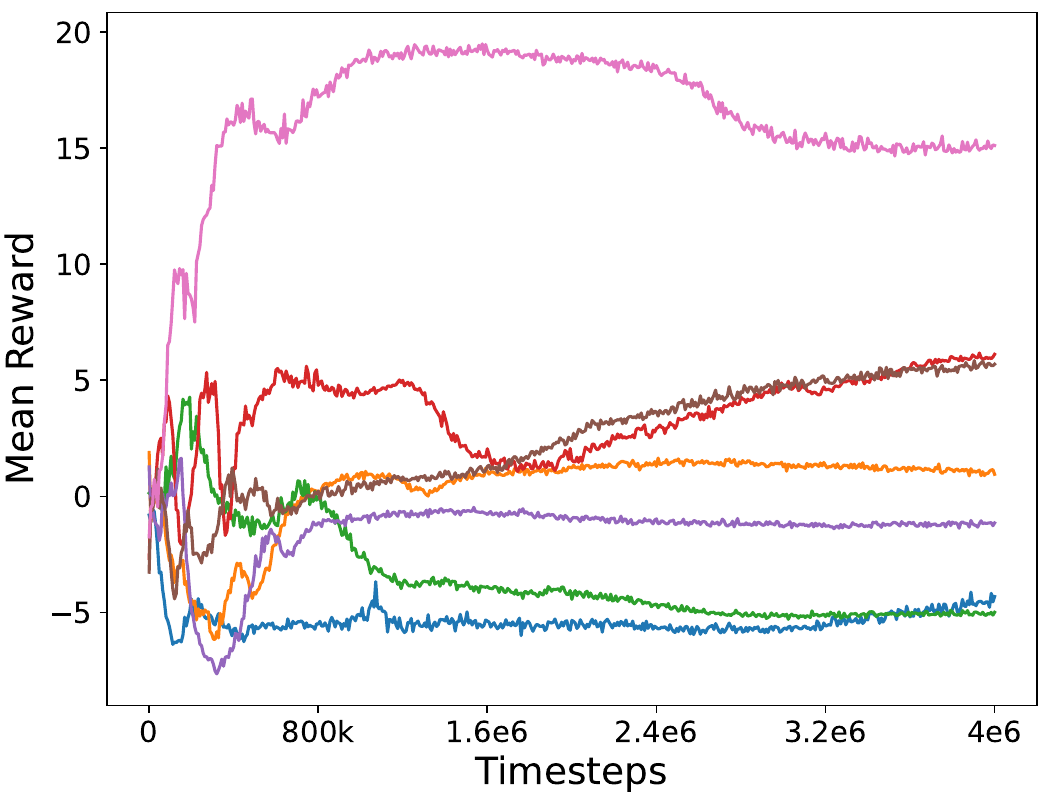}
         \label{fig:swimmer_rbrl_individual}
     }     
\caption{Results of individual runs in the human user study for the Cheetah and Swimmer environments (7 users in each case).}
\label{fig:human_study_results_individual}
\end{figure}

Figure~\ref{fig:human_study_results_individual} displays the individual reward curves for non-expert users who participated in the human user study, for the Cheetah and Swimmer tasks.

\newpage

\section{Modeling Rating Class Probabilities} \label{sec:rateprob}

\begin{figure}[ht]
\begin{center}
\centerline{\includegraphics[width=0.77\columnwidth]{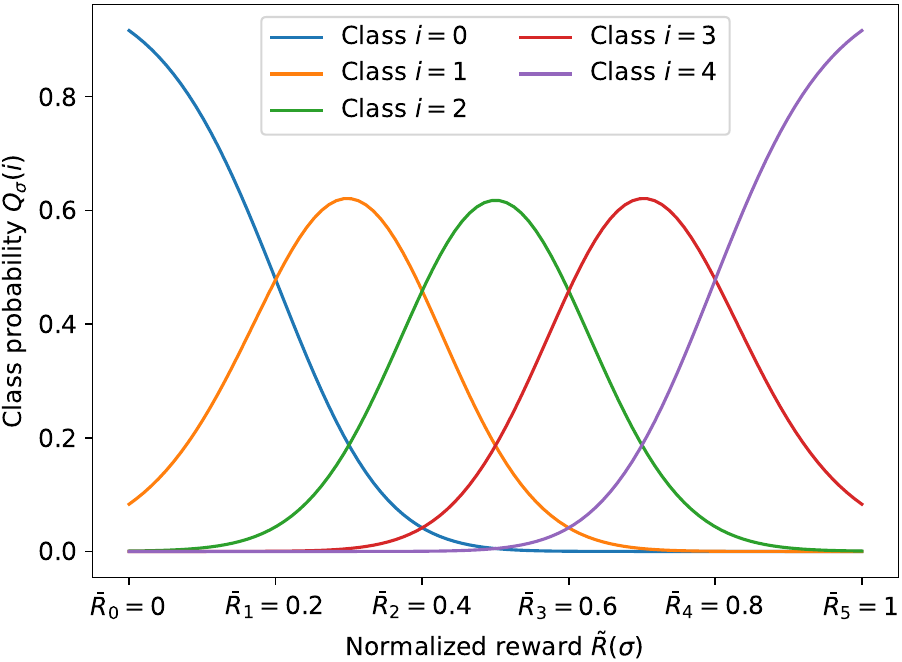}}
\caption{RbRL's modeled class probabilities $Q_\sigma(i)$. This figure illustrates Equation~\eqref{eq:Qfun_new}, evaluated with $n=5$ classes, $k=30$, and the rating category separation parameters hard-coded at $\{0, 0.2, 0.4, 0.6, 0.8, 1\}$. We can see that the probability of belonging to a class $i$ is maximized when the normalized return $\tilde{R}(\sigma)$ is halfway between $\bar{R}_i$ and $\bar{R}_{i + 1}$, and decreases as $\tilde{R}(\sigma)$ moves further from this point.}
\label{fig:rating_class_probs}
\end{center}
\end{figure}

Figure~\ref{fig:rating_class_probs} provides an intuitive illustration of the modeled class probabilities $Q_\sigma(i)$, as calculated via Equation~\eqref{eq:Qfun_new}, with specific relevant parameter choices detailed in the figure caption.

\section{Questionnaire}\label{sec:ques}

The three-part questionnaire given to the user study participants appears next.

\includepdf[pages=-]{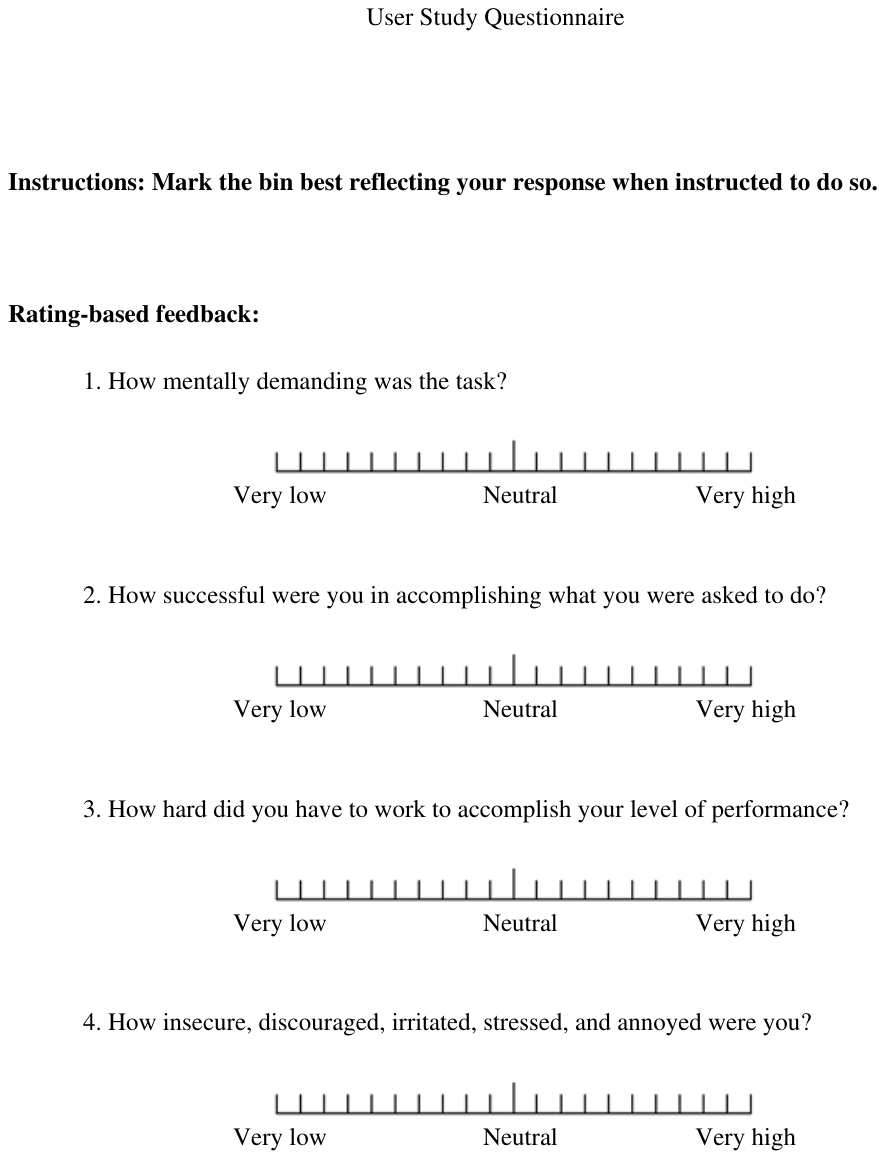}




\end{document}